\UseRawInputEncoding
\documentclass[journal]{IEEEtran}

\usepackage{times}
\usepackage{epsfig}
\usepackage{graphicx}
\usepackage{amsmath}
\usepackage{amssymb}

\usepackage{algorithm}
\usepackage{algorithmicx}
\usepackage{algpseudocode}
\usepackage{balance}
\usepackage{CJK}
\usepackage{multirow}
\usepackage{booktabs}
\usepackage{verbatim}
\usepackage{pifont}

\usepackage{acronym}
\usepackage[colorlinks,linkcolor=black, citecolor=black]{hyperref}

\newcommand{\vc}{{\bf c}}

\newcommand{\vf}{{\bf f}}

\newcommand{\vq}{{\bf q}}

\newcommand{\vx}{{\bf x}}

\newcommand{\vP}{{\bf P}}

\newcommand{\vX}{{\bf X}}

\begin{document}

\title{Hybrid Dynamic Contrast and Probability Distillation for Unsupervised Person Re-Id}

\author{De~Cheng,
    Jingyu~Zhou,
	Nannan~Wang$^{*}$~\IEEEmembership{Member,~IEEE,}
    Xinbo~Gao~\IEEEmembership{Senior Member,~IEEE.}
}
%



\maketitle

\begin{abstract}
Unsupervised person re-identification (Re-Id) has attracted increasing attention due to its practical application in the read-world video surveillance system. The traditional  unsupervised Re-Id are mostly based on the method alternating between clustering and fine-tuning with the classification or metric learning objectives on the grouped clusters. However, since person Re-Id is an open-set problem, the clustering based methods often leave out lots of outlier instances or group the instances into the wrong clusters, thus they can not make full use of the training samples as a whole. To solve these problems, we present the hybrid dynamic cluster contrast and probability distillation algorithm. It formulates the unsupervised Re-Id problem into an unified local-to-global dynamic contrastive learning and self-supervised probability distillation framework. Specifically, the proposed method can make the utmost of the self-supervised signals of all the clustered and un-clustered instances, from both the instances' self-contrastive level and the probability distillation respective, in the memory-based non-parametric manner. Besides, the proposed hybrid local-to-global contrastive learning can take full advantage of the informative and valuable training examples for effective and robust training. Extensive experiment results show that the proposed method achieves superior performances to state-of-the-art methods, under both the purely unsupervised and unsupervised domain adaptation experiment settings.

\end{abstract}

\begin{IEEEkeywords}
Unsupervised, Person Re-Id, Dynamic Contrastive Learning, Probability Distillation.
\end{IEEEkeywords}

\section{Introduction}

%
%
%
%
\IEEEPARstart{U}nsupervised person Re-Id aims to train a deep model capable of retrieving the person of interest from a large amount of unlabeled person Re-Id datasets. This task attracts increasing attention in the computer vision research field in recent years, due to the great demand in practical intelligent video surveillance and the expensive labeling cost with ever-increasing surveillance data. Thus, developing efficient and robust unsupervised person Re-Id system is very appealing not only in the academic area but also for the industrial field.  Existing unsupervised person Re-Id methods falls into the following two categories: 1) the purely unsupervised learning (USL) person Re-Id; 2) the unsupervised domain adaptation (UDA) person Re-Id. The purely USL methods don't use any labeled data for pre-training. They generally estimate the pseudo labels for the completely unlabeled training sample, then they gradually use the pseudo labels for classification or metric learning, the traditional method is named "Clustering and Fine-tuning" \cite{fan2018unsupervised}. The UDA methods usually adopt the two-stage training strategy, they first pre-train a model on the source labeled dataset, and then finetune the model on the unlabeled target datasets in an unsupervised manner. For the unsupervised fine-tuning stage, many state-of-the-art methods demonstrate the effectiveness of the pseudo-label-based USL methods~\cite{fu2019self,ge2020mutual,zhu2020self,zhang2019self,cheng2016person}. Besides, the UDA methods always exploit the relationships between the source and target datasets, and adapt the model trained on the source domain to capture the inter-sample relations in the target domain.
Generally the UDA methods are superior to that of the purely USL for person Re-Id, since the UDA methods could improve the performances on the target domain by transferring the learned knowledge from the labeled source domain.

In this paper, we formulate the USL and UDA methods into the same infrastructure. The proposed method can be easily extended from the purely USL to the UDA person Re-Id task, with effective performance improvements and requiring no complex training process.
Existing state-of-the-art USL methods integrate pseudo-label-based algorithm to the hybrid memory-based contrastive learning framework\cite{zhu2020self}, and the representative memory-based method first extract all the image features of the training data by the current feature extractor and store them in the memory bank. Then, they use a clustering method, like DBSCAN~\cite{ester1996density} or K-means~\cite{kanungo2002efficient}, to group the image features and generate the pseudo labels, thus the cluster ID is assigned to each image as its person identity. Finally, the deep model is trained with the contrastive loss or some other identity classification loss using the corresponding image features contained in the memory bank. Such memory-based methods have the following merits:1) The image features can be updated gradually as a whole, then the clustering step can be processed online and the estimated pseudo label of the image could be gradually corrected in the training process; 2) The memory-based methods can utilize large batch sizes of the training data in every training step. This could great benefit the contrastive learning, where the more sample pairs are used in each training batch, the better performance improvements can be obtained, as larger batches are likely to contain more informative examples. Thus, this memory-based method can overcome the limitations of the GPU memory in the training process, then the deep model can be optimized globally in each training step.

Although these memory-based contrastive learning methods have achieved remarkable performance, there still contains a large gap between the purely USL and supervised methods. After detailed analysis of these methods, we consider that there exist the following two major limitations hindering the further performance improvements for the purely USL person Re-Id: 1) The traditional "Clustering-Finetuning"-based methods~\cite{fan2018unsupervised} didn't explore the valuable training samples from different viewpoints or cameras with the same identity for person Re-Id tasks. Since the unsupervised person Re-Id task is a little different from the traditional unsupervised classification task, the true cluster for the person Re-Id task always contains samples from different viewpoints or different cameras, which are the most informative training examples but make the intra-class distances always larger than that of the inter-class distances. Especially in the clustering process, this leads the samples in each cluster only contains samples from the same viewpoint or camera. Some methods may relax the cluster conditions, usually it could only get a small portion of samples in different viewpoints with the same identity, or sometimes get the incorrect samples. However, the sample pairs in different viewpoints or cameras with the same identity are the most valuable training samples for the person Re-Id task. 2) The traditional USL methods did not make full use of the training samples, especially for lots of un-clustered instances. Usually, the clustering method group the training data into the confident clusters and some un-clustered outlier instances. To ensure the reliability of the generated pseudo labels, recent existing methods~\cite{ge2020structured,zhang2019self,fu2019self} simply discard the un-clustered outliers from being used for training, and just use the clustered instances to generate pseudo labels for supervised learning. However, the un-clustered outliers always contains many hard training examples, which might actually be very difficult but valuable for the person Re-Id task. Just simply abandoning such training examples might critically hurt the recognition performances.

To address the above mentioned problems, we propose the hybrid dynamic contrastive and probability distillation framework for the purely USL and UDA person Re-Id. For the instances in each cluster, we dynamically  find its corresponding hardest positive and negative pairs from the global memory-bank level and the local mini-batch level. Mining hardest examples for each instance in the current training batch from the global memory bank, can help us use the most informative and valuable examples to optimize the deep model. However, such hardest examples may contain some incorrect training pairs, because the clustering results is usually not accurate in some cases. In order to keep moderate model updating, we also consider the local batch-level contrastive learning. Since the instances in each mini-batch are randomly sampled from the whole training dataset, and the number of clusters is relatively very small, thus it could have relatively low probability to select the incorrect negative/positive sample pairs for the contrastive learning. By combining this local-to-global contrastive learning with dynamic hardest sample mining, the model optimization process could be very stable while effective and robust.

Specially, we also propose the instance-level self-contrastive and the probability distillation method, to make full use of the self-supervised signals of all the clustered and un-clustered instances in the non-parametric manner. We consider the self-supervisor signals from the following two perspectives: 1) We treat the un-clustered outlier instances as independent classes~\cite{zhu2020self}, and then  maximize the distance between the similarity of the instance with its corresponding strong augmented instance, and the similarity of the instance with its nearest negative examples in each clusters; 2) The proposed the probability distillation framework, relies on the assumption that the model should output very similar probability predictions when fed perturbed augmentation of the same image. The factor that contribute most to the success of the proposed algorithm is that we have cast the unsupervised person Re-Id task as the gradual semi-supervised learning. Given the previously clustered instances stored in the memory bank, we creatively obtain the probability distribution of each instance on the clustered centroids, in a non-parametric manner. More specially, on top of the "Mean-Teacher"\cite{tarvainen2017mean} framework, we use the weak and stochastic data augmentation of the image as the input of the teacher channel, and the strong augmentation of the same image as the input of the student channel. As the parameters of the teacher model is updated by the Exponential Moving Average(EMA) of the student model, this will provide a more stable target and was empirically found to be very effective. Besides, after obtaining the probability distribution of each instance from the teacher model, we further transform the probability into the near "one-hot" output by using the sharpen technics~\cite{li2020dividemix}. Finally, the proposed probability regression technic is used between the probability output of the student model and the sharpened probability of the teacher model. Thus, the proposed method can effectively utilize all the self-supervised signals in the training dataset.

Our contributions are summarized as three-fold:
\begin{itemize}

\item Besides utilizing the clustering-based pseudo labels for USL person Re-Id, the proposed method can make full use of the self-supervised signals of all the clustered and un-clustered instances, from both the instances' self-contrastive level and the probability distillation perspective in the memory-based non-parametric manner.

\item The proposed hybrid local-to-global contrastive learning with dynamic sample mining, can take full advantages of the informative and valuable training examples for person Re-Id task, which also makes the model optimization process very stable while effective and robust.


\item The proposed method significantly outperforms most of state-of-the-art algorithms on multiple person Re-Id datasets, under both the purely USL and UDA experiment settings. Specially, our method outperforms the state-of-the-art method~\cite{zhu2020self} by 8.6\%, 3.7\%, 5.5\%, 11.0\% in terms of mAP on Market1501,Duke, MSMT17 and PersonX datasets.
\end{itemize}

\section{Related Works}
In this section, we review the most relevant works with ours in the following perspectives: 1)Deep unsupervised person Re-Id, which includes both the purely unsupervised person Re-Id and the unsupervised domain adaptation person Re-Id; 2)The general unsupervised or self-supervised representation learning; 3) Memory-based deep metric learning.

\textbf{Deep unsupervised person Re-Id} methods can be summarized into the purely USL person Re-Id and the UDA Re-Id. Generally, the recent effective purely USL person Re-Id are clustering-based methods,  which generate hard/soft pseudo labels, and then finetune/train the deep models based on the pseudo labels~\cite{fan2018unsupervised}. The pseudo labels can be obtained by clustering the sample features or measuring the similarities among the instances' features. The representatives of such pseudo-label-based methods include \cite{song2020unsupervised,fu2019self,zhu2020self}, which have got state-of-the-art performances at present. However, such pseudo-label-based method highly depends on the precision of the estimated pseudo labels, since the noisy labels could degrade the model performances. To improve the accuracy of the estimated pseudo labels and mitigate the effects caused by the incorrect labels, many improved pseudo-label estimation methods were proposed, such as SSG\cite{fu2019self} adopted human local features to assign multi-scale pseudo labels, PAST~\cite{zhang2019self} try to utilize multiple regularization to overcome this problem, and a lot of methods proposed some reliability evaluation criteria to further modify the generated pseudo labels~\cite{ge2020mutual}. Besides estimating the hard pseudo labels, some researchers proposed the soft-label and multi-label based USL methods for person Re-Id. For instance, Yu etal~\cite{yu2019unsupervised} proposed the soft-label based method by measuring the similarities between the current person image with the reference clustered images, and Wang etal~\cite{wang2020unsupervised} transferred the USL person Re-Id into the multi-label classification problem, then it gradually found the true labels with the help of some label estimation strategy and consistency regularization, to improve the precision of the estimated pseudo labels. Recently, some methods introduce mutual learning among two/three collaborative networks to mutually exploit the refined soft pseudo labels of the peer networks as supervision~\cite{ge2020mutual}. The above mentioned two types of pseudo-label-based methods are widely used in both of the purely USL and UDA person Re-Id tasks. While for the UDA person Re-Id, some domain translation-based methods are specially proposed, because the source domain labeled data can be used in this task. The representative work include the style-GAN~\cite{wei2018person} based translation methods from the source domain to the target domain images, which transform the images from the source domain to match the image styles in the target domain while keep the person identities appeared in the source domain. These methods transfer the purely USL task to the semi-supervised task, then the previously mentioned purely USL methods can be fully used. Besides, some other methods exploits the valuable information across the source and target domains, then explore some underlying relationships among them and finally construct the joint learning framework by using the training data in both the source and target domain. The proposed method in this paper falls into the category of purely USL, but we have also extended the method to the UDA task by fully exploits the useful training examples in the source domain without consuming too much training resources. We have considered the local and global label estimation quality to make the training procedure stable and efficient.

\textbf{Unsupervised or Self-supervised Representation Learning}  aims to learn the feature representations from the totally unlabeled data. Besides the above mentioned clustering pseudo-label-based methods, the unsupervised representation learning have another two categories: the generative models and the self-supervised learning. In this paper, we just talk about the relevant self-supervised learning methods. Early self-supervised learning methods in computer vision were based on the proxy tasks, such as the solving jigsaw puzzels~\cite{noroozi2016unsupervised}, colorization and rotation prediction methods~\cite{zhang2016colorful,doersch2015unsupervised}. Recently, some contrastive learning based self training methods were proposed and achieved promising results, the representative methods includes MoCo~\cite{he2020momentum}, SimCLR~\cite{chen2020big} and BYOL~\cite{grill2020bootstrap}. The idea behind such self-supervised learning methods is that the same image in different views or augmented by different methods should have consistent outputs when processed by the same deep model. There also exist many other works aiming to improve previous self-supervised learning approaches from different perspectives~\cite{he2020momentum}. The most relevant work to ours in the self-supervised learning research area is the MoCo~\cite{he2020momentum}, BYOL~\cite{grill2020bootstrap} and MeanTeacher~\cite{cai2021exponential}, where the Moco method builds a dynamic dictionary based contrastive learning framework with a moving averaged encoder, and it narrows the gap between the supervised and unsupervised learning in some domains. The BYOL method further improves this method and achieved better results than previous work ~\cite{chen2021exploring,grill2020bootstrap} by optimizing a feature regression loss. Both of these two methods borrow the ideas from the mean-teacher approach including the student-teacher framework and the EMA parameter updating strategy. Our proposed method falls into this category, but we are different with them. The proposed method fist transform the unsupervised person Re-Id into the semi-supervised problem, then we creatively transform the instance feature into the class probability, and finally the probability regression objective is used to further improve the performances, instead of the instance feature itself. To some extent, such strategy can mimic the traditional combination of identity loss and the contrastive loss, which have got superior performances on the supervised person Re-Id task.

\textbf{The memory dictionary based deep metric learning } methods present promising results on the unsupervised visual representation tasks. Besides that, state-of-the-art unsupervised person Re-Id methods also build memory dictionary for the contrastive learning, and many strategies have been proposed to consistently update the memory dictionary which makes the deep metric learning very effective. Especially for the USL/UDA person Re-Id, Ge etal~\cite{zhu2020self} proposed the self-paced contrastive learning for UDA person Re-Id tasks. Zheng etal~\cite{zheng2020exploiting} also proposed to exploit the sample uncertainty for UDA person Re-Id task while affiliating with the memory based contrastive learning. These methods can leverage the memory bank to measure the similarity between a sample and the instances stored in the memory, which helps to mine hard negative examples across batches and increase the contrastive power with more negatives, and finally better train the deep model. We introduce the local-to-global memory bank contrastive learning to jointly utilize the estimated pseudo labels and the self-supervised consistency regularization. We further validate the synergy between the pseudo label estimation and the self-supervised probability regression module, where the probability regression module can be interpreted as a form of knowledge distillation with pseudo labels or the approximately projected labels.
\begin{figure*}[!t]
\centerline{\includegraphics[width=18cm,height=8cm]{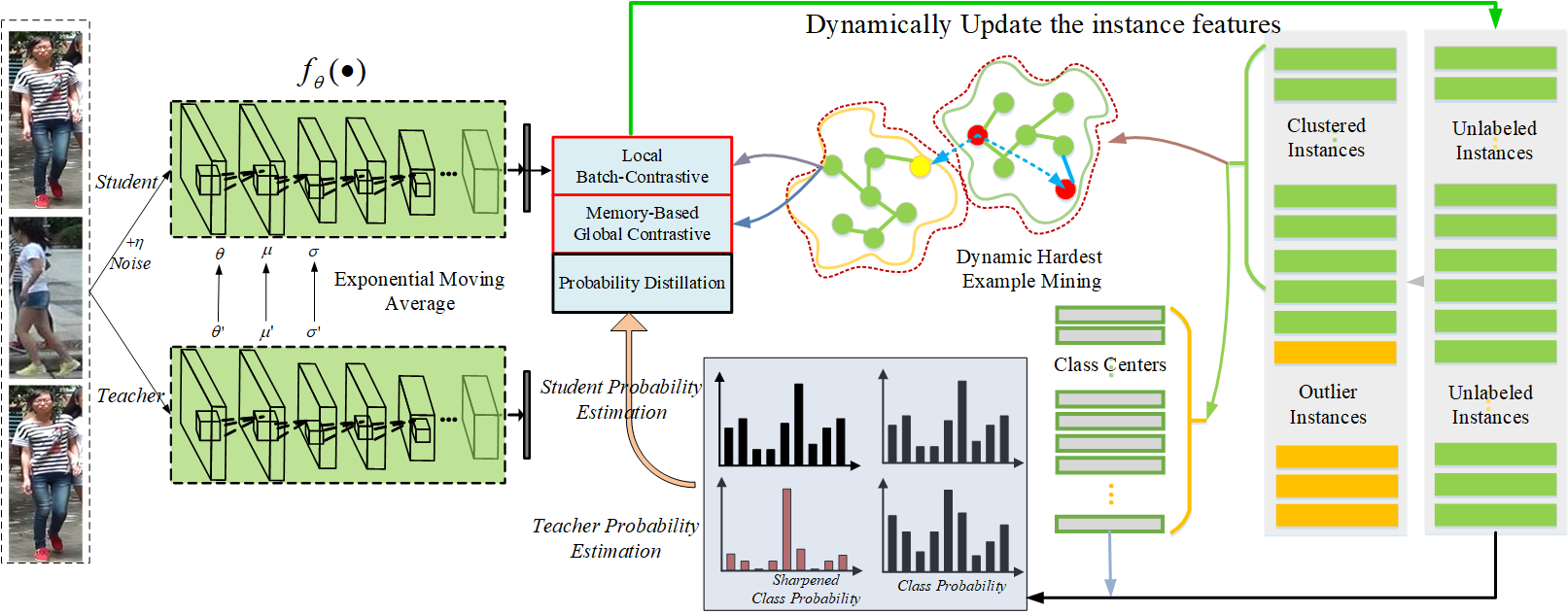}}
\caption{This is the proposed framework for unsupervised person Re-Id. It mainly consists of two modules: the hybrid dynamic contrastive learning and the probability distillation component. For the memory-based hybrid dynamic contrastive learning, it first uses the clustering method "DBSCAN" to group the unlabeled instances into the clustered instances and the outlier instances, then we use the memory-based local-to-global contrastive learning objectives with dynamic hardest example mining strategy to optimize the deep model. The instance features stored in the memory bank is updated dynamically in the training process. For the probability distillation module, the training images with different data augmentation strategies are processed through the student and teacher network, then we project the image features from the student and teacher network onto the pseudo class centers to get their class responses. Followed by the sharpen technics, we finally minimize the probability regression objective to make full use of the self-supervised signals of the training data.}
\label{overallFramework}
\end{figure*}

\section{Method}
To tackle the challenges on the USL person Re-Id task, we propose the hybrid dynamic contrast and probability distillation framework as shown in Figure~\ref{overallFramework}. It mainly consists of the following two modules: 1) The self-supervised framework with probability distillation;2) The hybrid dynamic local-to-global contrastive learning. The above two modules are both optimized in the global memory bank. In Figure~\ref{overallFramework}, it demonstrates the pipeline of the proposed method. Specially, we also use the Resnet50~\cite{he2016deep} as the backbone network to extract the semantic features of the input image. Similar to~\cite{zhu2020self}, the DBSCAN~\cite{ester1996density} clustering method is used to group the instance features into several clusters and the remaining un-clustered individual instances, then we assign the cluster ID to each instance as its pseudo label. Note that there are two important hyper-parameters in the DBSCAN~\cite{ester1996density} algorithm, one is the maximum distance $\alpha$ between the examples to be considered as the neighborhood of each other, and another is the minimum number of samples in a neighborhood for an instance to be considered as the center instance. In this paper, we experimentally set these two hyper-parameters as 0.5 and 5 respectively according to SPCL method~\cite{zhu2020self}. For the un-clustered instances, we just simply consider them  as the individual classes. When the pseudo labels are generated,  the dynamic local-to-global contrastive learning is performed in the training stage. During optimizing the contrastive learning objectives, the dynamic hardest example mining strategy is performed in the training batch level and the memory based cluster level. The instance features stored in the memory are dynamically updated using the mined hardest training example during the training process. The local-to-global contrastive learning process can help us not only utilize the most valuable and informative training examples in the cluster level among the whole training dataset, but also make the model optimization process avoid training error amplification caused by the noisy clusters, thus keep the training process very stable while effective and robust.
Meanwhile, to make full use of all the self-supervised signals of all the clustered and un-clustered instances, we propose the instance level self-contrastive learning on the un-clustered individual training samples and the probability distillation objectives on the whole training data, in the non-parametric manner. As shown in the left side of  Figure~\ref{overallFramework}, each training image augmented with different strategies is processed through the student and teacher networks, then we project the image features from the student and teacher network onto the cluster centroids to get
their class predictions, where the cluster centroids is dynamically computed by the instances' features stored in the memory bank. Followed by the probability sharpen technics, we finally minimize the probability regression objectives. We formulate the above cluster contrast and the probability distillation modules into the unified learning framework.

Noted that, the training scheme alternates between the following two steps: 1) Grouping the un-labeled training examples into clusters and un-clustered instances by using the DBSCAN method~\cite{ester1996density} with the instances stored in the memory bank; 2) optimizing the feature extractor with the proposed local-to-global contrastive learning and the self-supervised probability distillation objectives, and dynamically update the hybrid memory bank with instance features extracted by the teacher network at the sametime.




\subsection{Hybrid Local-to-Global Cluster Contrast Learning}
As we all know, the most valuable and informative training examples for person Re-Id task, are the examples which are captured from different cameras or views of the same identity. But for the USL person Re-Id task, the primary grouped clusters usually contain examples mainly from one camera view, because the intra-class distance from different camera views of the same identity is usually larger than that of the inter-class distance from the same camera view of different identities, due to the illumination and resolution changes. This is the main difference between the USL person Re-Id and the traditional unsupervised classification task, where the traditional unsupervised classification tasks usually process the training data which are uniformly distributed. Here, we introduce the memory-based local-to-global cluster contrast method with hardest data mining strategy to overcome this challenge in some respects.  To make the description of the proposed method readable, we keep some similar definitions as described in~\cite{zhu2020self}, and demonstrate the algorithm from the following two aspects£º1) the local-to-global cluster contrastive learning; 2) The memory bank initialization and updating strategy.

Given the training datasets $\vX$ without any ground-truth label, we initially use the clustering method DBSCAN~\cite{ester1996density} to group the training data into clusters $\vX_c$ and the un-clustered individual instances $\vX_o$. Then we assign the cluster ID to each instance as the pseudo label, and here we just consider the un-clustered instances as individual clusters. In the proposed probability distillation framework, the parameters of the student network is optimized using these training examples. Given the training sample $\vx_i$, we define $\vf_\theta(\vx_i)$ as its feature representation vector, abbreviated as $\vf_i=\vf_\theta(\vx_i)$, $\vx_i\in{\vX_o\cup \vX_c}$. The memory based global contrastive loss $\mathcal{L}_{\text{GMemory}}$ can be defined as the following Eq~\ref{e:GCont},

\begin{equation}
     -\log \frac{ \mathbf{exp}<{\vq_i, \vf^+)/\tau}> }{ \sum\limits_{k=1}^{N_c} \mathbf{exp}{\left(<\vq_i, \vf_{ck^*}>/\tau\right)} + \sum\limits_{k=1}^{N_o} \mathbf{exp}{\left(<\vq_i, \vf_{k}>/\tau\right)} },
     \label{e:GCont}
\end{equation}
where $\vq_i$ indicates the $i-th$ query feature in current mini-batch,  $\vf^+$ indicates the positive class prototype corresponding to the $i-th$ query feature $\vq_i$, $\tau$ is the temperature hyper-parameter which is empirically set to 0.05 in our experiments, $<.,.>$ denotes the inner product between two feature vectors to measure the similarity, $N_c$ is the number of grouped clusters in current memory bank, $N_o$ is
the number of un-clustered individual instances in current memory bank. If the current query feature $\vq_i$ falls into the  grouped clusters, $\vf^+$ is the hardest positive example for $\vq_i$ in its corresponding cluster, where the "hardest positive" means the selected example is the farthest to $\vq_i$ in the same cluster. Thus, $\vf_{ck*},c\in\{1,...,N_c\}$ contain one hardest positive example for the $\vq_i$'s corresponding cluster, and the rest $N_c-1$ hardest negative examples in the corresponding negative clusters in the memory bank, where the "hardest negative" means the nearest negative example to the query example $\vq_i$ in each cluster, and $\vf_k, k\in\{1,...,N_o\}$ is the un-clustered instance features in the memory bank. If $\vq_i$ falls into the un-clustered individual groups, we would set $\vf^+=\vf_k$ as the outlier instance feature corresponding to $\vq_i$ in the memory bank, and $\vf_{ck*}$ is the hardest negative examples in all the $N_c$ clusters in the memory bank corresponding to $\vq_i$. Thus, in this situation, the proposed method can make full use of the self-supervised signals from the instance-level self-contrastive learning.  Please note that $\vq$ represents the features in current mini-batch, and $\vf$ represents the features stored in the memory bank in Eq~\ref{e:GCont} and the following descriptions.

Above we have described the proposed memory-based global cluster contrastive loss. In order to keep the learning procedure much more stable and moderate, we further propose the local batch contrastive loss $\mathcal{L}_{\text{LBatch}}$ to facilitate the memory-based global contrast loss as follows,
\begin{equation}
     -\log \frac{ \mathbf{exp}{<\vq_i, \vq_i^+>/\tau} }{\mathbf{exp}{\left(<\vq_i, \vq_i^+>/\tau\right)} + \sum\limits_{\vq_j \in B, y_i\neq y_j} \mathbf{exp}{\left(<\vq_i, \vq_j>/\tau\right)} },
     \label{e:LCont}
\end{equation}
where $y_i$ and $y_j$ denote the pseudo labels for the examples $\vq_i$ and $\vq_j$ in current mini-batch, $B$ is the batch size. we can clearly see that the local batch contrastive loss $\mathcal{L}_{\text{LBatch}}$ is the traditional contrastive loss, where the hardest positive and negative training pairs are selected in current min-batch.

The memory-based global hardest contrast loss $\mathcal{L}_{\text{GMemory}}$ explores to utilize the most informative and valuable examples from the global memory bank to optimize the deep model, but such hardest examples usually contain some incorrect training pairs since the unsupervised clustering results is not accurate in some cases. In order to keep moderate model updating, we also consider the local batch-level contrastive loss $\mathcal{L}_{\text{LBatch}}$, which could have relatively low probability to select the incorrect negative example pairs, since the number of clusters in each mini-batch is relatively small. Besides, leveraging the training pairs in the current mini-batch can utilize the latest instance features to update the deep model. Finally, the hybrid local-to-global contrastive loss with hardest data mining strategy $\mathcal{L}_{\text{L2G}}$ constitutes as follows,

\begin{equation}
     \mathcal{L}_{\text{L2G}} = \mathcal{L}_{\text{GMemory}} + \mathcal{L}_{\text{LBatch}}.
     \label{e:HybridLoss}
\end{equation}

\subsection{The memory bank and Its updating strategy}
We store all the instances' feature $\{\vf_1,\vf_2,...,\vf_N\}$ in the memory bank. Here $N$ stands for the number of the whole training examples in the dataset. At the beginning, we use the deep CNN model pre-trained on the ImageNet to extract the instance features, then we use the DBSCAN~\cite{ester1996density} method to cluster the instances into several clusters as initialization. The clustering results would impact the learned representations. If the clustering is perfect, merging the instances to its true clusters would no doubt improve the performance, which is the upper bound or ideal conditions. However, in practice, the unsupervised learning usually would merge some instances into the wrong clusters inevitably, which could do harm to the final performance.
For clustering-based unsupervised person Re-Id, the accuracy of clustering results would gradually improves as the model updates, thus better feature representations can be learned finally.
In the proposed algorithm, we first group all training instances into the clusteres and the un-clustered individual instances, to obtain the pseudo labels of the instances with the cluster ID.
Following, we optimize the proposed learning objectives with these pseudo labels for $T$ epoch on the training dataset, here $T=2$  in our experiments. As the model parameters update, the instances' features stored in the memory bank would also be updated in the training process. After every $T$ epoch, we use the clustering method "DBSCAN" to generate the pseudo labels for all the training instances in the memory bank.

Please note that, the instances' features stored in the memory bank are updated at each training iteration dynamically, instead of updating them totally using current model parameters every $T$ epoch. Experiments shows that updating the instances' feature vectors dynamically is better than that of updating them as a whole every several iterations. In order to make the contrastive loss work well in the training process, we sample $P$ person identities and a fixed number of $H$ instances for each person identity using the obtained pseudo labels in each training batch. As a consequence, there are $P\times H$ query instances in each mini batch~\cite{dai2021cluster}. The same as previous instance-level feature memory methods~\cite{dai2021cluster}, we also update all the $P\times H$ query instances' features in the memory. The difference is that we use the instance features extracted from the teacher model in the framework to update the instance features stored in the memory bank, as the teacher model gradually grows better than the student model in the training process. In each iteration, the $P\times H$ extracted feature vectors in current mini-batch are utilized to update the corresponding feature vectors stored in the memory bank as follows,
\begin{equation}
     \vf_i\leftarrow m\vf_i+(1-m)\vq_i,
     \label{e:Update}
\end{equation}
where $m\in[0,1]$ is the momentum coefficient for updating the instance features in the memory bank and is set to 0.3 in our experiments. $\vq_i$ is the instance feature vector in current mini-batch, and $\vf_i$ is the feature vector stored in the memory bank.

\begin{figure}[!t]
\centerline{\includegraphics[width=8.2cm]{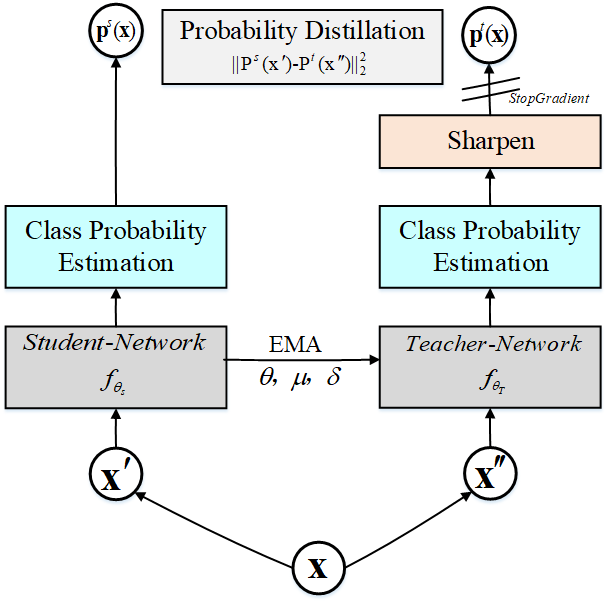}}
\caption{This is the proposed probability distillation framework. Two augmented views of one input image $\vx$ are processed by the same network architecture named "Student-Network" and "Teacher-Network" respectively, both of which contains the backbone network (Resnet50) and the class probability estimation module. Then the sharpen technic and stop-gradient operation are applied on the teacher network, where the parameter in the teacher network is updated by EMA method. The model use the L2 regression loss to maximize the similarity of the estimated probability between the student and teacher networks.}
\label{fig3}
\end{figure}

\subsection{Self-Supervised Probability Distillation}
Apart from learning discriminative image features based on the pseudo labels generated by the clustering methods, our method also considers the self-supervised signals through the proposed pseudo probability distillation framework, which can also be regarded as a sort of knowledge distillation approach.

The proposed probability distillation framework is a learning paradigm which trains a student neural network $\vf_{\theta_s}(\cdot)$ and a teacher network $\vf_{\theta_t}(\cdot)$. The teacher and student networks share the same network architecture, but they are parameterized by $\theta_s$ and $\theta_t$ respectively. Give each input image $\vx$, our network takes two randomly augmented views $\vx'$ and $\vx''$ as the input for the student and teacher network, respectively. These two augmented views are processed by the encoder network $\vf_{\theta}(\cdot)$ and the class probability projection module, where the encoder network is formed by a backbone network Resnet50~\cite{he2016deep}. For the pseudo label based unsupervised learning, we first group all the training instances into the clusters and un-clustered outliers. Then, we can obtain the cluster centroid based on the grouped instances' features. Thus we can compute the probability distribution of each instance belonging to the corresponding cluster centers. Specifically, we compute the cluster center with the mean feature vectors of each cluster set from $\{\vf_i\}$ in the memory bank, as illustrated in Eq~\ref{e:Center},
\begin{equation}
     \vc_k=\frac{1}{|\mathcal{M}_k|}\sum_{\vf_i\in \mathcal{M}_k}\vf_i,
     \label{e:Center}
\end{equation}
where $\mathcal{M}_k$ demotes the $k$-th cluster set that contain all the feature vectors within the $k$-th cluster in current memory bank $\mathcal{M}$, and $|\cdot|$ denotes the number of instances in the subset, $\vf_i$ is the feature vector stored in the memory bank. Please note that only the grouped clusters are used to computer the cluster centers.

Given each query image $\vx_i$, both networks output their corresponding feature representations $\vq_i^s$ and $\vq_i^t$, then we compute their probability distributions on the obtained cluster centers. For the feature vector $\vq_i^s$, the corresponding probability distribution $\vP^{s}(\vx_i)$ is illustrated in the following Eq~\ref{e:Probability},
\begin{equation}
     \vP_{k}^s(\vx_i) = \frac{\mathbf{exp}(<\vq_i^s, \vc_k>/\tau_s)}{\sum_{k=1}^{K}\mathbf{exp}(<\vq_i^s, \vc_k>/\tau_s)},
     \label{e:Probability}
\end{equation}
where $k\in{1,...,K}$, and $K$ is the number of grouped clusters at present, please note that $K$ is changing in the whole training procedure, because the unsupervised person ReId is an openset problem, we don't know the true number of IDs previously, and we cluster the whole instances every several epoches using the DBSCAN methods in the algorithm. Then for the clustered instances, the obtained probability distribution $\vP_{k}(\vx_i)$ could represent the probability that the instance $\vx_i$ belongs to the $k$-th cluster. While for some instances in the rest un-clustered outlier subset, the probability distribution can only be considered as the feature projection on the obtained cluster centers, because some outliers maybe don't belong to any of the previously obtained clusters. This is also one of the major differences between the unsupervised classification and the unsupervised open-set problems, such as person Re-Id. In Eq~\ref{e:Probability}, $\tau_s>0$ is the temperature parameter that controls the sharpness of the output distribution. Similar formula holds for the teacher network $\vf_{\theta_t}(\cdot)$, and we obtain $\vP^t(\vx_i)$ for the instance feature vector $\vq_i^t$ with temperature parameter $\tau_t$.

The proposed self-supervised probability distillation module works as a regularisation term between these two networks' output probability distributions. Given a fixed teacher network $\vf_{\theta_t}(\cdot)$, we learn to match these probability distributions by minimizing the Mean Square Loss $\mathcal{L}_{\text{S}}$ in the self-supervised manner, as illustrated in Eq.~\ref{e:Ls},
\begin{equation}
     \mathcal{L}_{\text{S}} = ||\vP^s(\vx_i) - \vP^t(\vx_i)||_2^2.
     \label{e:Ls}
\end{equation}
Note that in Eq~\ref{e:Ls}, the probability distribution $\vP^t(\vx_i)$ from the teacher network is fixed. Minimizing the loss function $\mathcal{L}_{\text{S}}$ could directly optimize the parameters of the student network $\vf_{\theta_s}(\cdot)$, while the parameters  in the teacher network $\vf_{\theta_t}(\cdot)$ is updated by Exponential Moving Average(EMA) methods based on current parameters in the student network $\vf_{\theta_s}(\cdot)$. Unlike the traditional knowledge distillation, we don't have a predefined high quality model as the fixed teacher model, we build it from the past iteration of the student network. The parameter update rule "EMA" is $\theta_t\leftarrow\lambda\theta_t + (1-\lambda)\theta_s$, and $\lambda$ is a smoothing coefficient hyper-parameter. In our algorithm, we not only update the parameter $\theta_t$ in the teacher network $\vf_{\theta_t}(\cdot)$ with the parameter $\theta_s$ in the student network $\vf_{\theta_s}(\cdot)$, but also the parameters $\mu_t$ and $\sigma_t$ in the BatchNorm layer of the teacher network are updated using the EMA method with the parameters $\mu_s$ and $\sigma_s$ in the BatchNorm layer of the student network~\cite{cai2021exponential}, $\mu_t\leftarrow\lambda\mu_t + (1-\lambda)\mu_s$ and $\sigma_t\leftarrow\lambda\sigma_t + (1-\lambda)\sigma_s$.

\begin{table*}[]
	\footnotesize
	\centering
	\caption{The statistics of the four datasets used for training and evaluations in our paper.}
	\setlength\tabcolsep{10pt}
	\vspace{10pt}
	\scalebox{1.0}[1.0]{
		\begin{tabular}{l|cccccc}
			\toprule
			Dataset& $\sharp$ train IDs & $\sharp$ train images & $\sharp$ test IDs &$\sharp$ query images &$\sharp$ cameras &$\sharp$ total images \\
			\midrule
			Market-1501 &751 &12936 &750 &3368 &6 &32217 \\
			MSMT-v2 &1041 &32621 &3060 &11659 &15 &126441 \\
            DukeMTMC &702 &16522 &702 &2228 &8 &36442 \\
            PersonX &410 &9840 &856 &5136 &6 &45792 \\
			\bottomrule
	\end{tabular}}
	\label{table:datasets}
\end{table*}

\subsection{The Final Objective Function}
Considering the above descriptions, the final objective function in our algorithm consists of the pseudo label based local-to-global contrastive loss and the self-supervised probability regression, as denoted in Eq.~\ref{e:TotalLoss},
\begin{equation}
     \mathcal{L}_{\text{total}} = \mathcal{L}_{\text{L2G}} + \gamma \mathcal{L}_{\text{S}},
     \label{e:TotalLoss}
\end{equation}
where $\gamma$ is the hyper-parameter to balance these two objectives. By optimizing Eq.~\ref{e:TotalLoss}, the proposed algorithm can not only well utilize the estimated pseudo labels, but also can make full use of the self-supervised signal from both the instances¡¯ self-contrastive level and the probability distillation perspectives in the memory-based non-parametric manner, to learn the discriminative image features for the unsupervised person Re-Id task.

\section{Experiments}\label{SEC:experiments}

\subsection{Datasets and Evaluation Protocal}\label{SEC:datasets}
We evaluate the proposed method on four large-scale person re-Id benchmark datasets, namely: Martket1501~\cite{zheng2015scalable}, DukeMTMC~\cite{ristani2016performance}, MSMT17-v2~\cite{wei2018person}, and PersonX~\cite{sun2019dissecting}. Among the four benchmark datasets, PersonX~\cite{sun2019dissecting} is one synthetic dataset, it contains the manually designed obstacles such as the random erasing, different resolution and lighting conditions. The rest three datasets are widely used real-world person re-Id datasets. We also illustrate the statistics of the four datasets used for training and evaluation in Table~\ref{table:datasets}.

Following existing person Re-Id methods~\cite{zheng2015scalable}, we adopt the mean Average Precision (mAP) and the Cumulative Matching Characteristic (CMC) as the evaluation metrics, where we report Top-1, Top5 and Top-10 of the CMC metric in this paper. We don't adopt any post-processing technics, such as re-ranking~\cite{zhong2017re} or multi-query fusion methods~\cite{zheng2015scalable} for the experiment evaluation.  In the experiment settings, besides the traditional purely USL person Re-Id which doesn't use any labeled data in the whole training procedure, we also adopt the UDA in the experiment to illustrate the effectiveness of the proposed method. Specifically, we use these two real-world datasets Market1501 and DukeMTMC to work as the source and target dataset alternatively, and finally report the matching performance under this UDA experiment settings.

\subsection{Implementation Details}
The backbone network we used to extract the image features is the Resnet50~\cite{he2016deep}, we remove the sub-modules after the ``conv5$\_$x"~\cite{he2016deep} and add the global average pooling (GAP) followed by the batch normalization and L2 normalization layers, finally it will produce 2048-dimensional features for each image. The backbone network is initialized with the parameters pre-trained on ImageNet~\cite{deng2009imagenet}. At the beginning, we just use the model pre-trained on ImageNet to extract the features of all the training examples for initialization, then the DBSCAN~\cite{ester1996density} method is used for clustering to generate the pseudo labels on all the training datasets.

In the training stage, the input images are resized to $256\times 128$ on all the datasets, the batchsize is set to 256. For each mini-batch, we randomly select 16 pseudo identities and each identity has 16 different images. For the two-stream probability distillation networks, we use different data augmentation strategies for each input branch, such as random cropping, horizontal flipping and random easing. The parameters $\tau$ in Eq.~\ref{e:LCont} and Eq.~\ref{e:GCont} are all set to 0.6, and the parameter $\tau_s$ in Eq.~\ref{e:Probability} is set to 1.0 for the student model and 0.5 for the teacher model. The parameter of the maximum neighbour distance for DBSCAN~\cite{ester1996density} is set to 0.5. We adopt Adam as the optimizer with learning rate setting as $3.5e-4$, and step-size is 30 which means that the learning rate degrades $1/10$ every 30 epoches, and the weight decay is set to $5e-4$, the parameter of momentum is 0.2. In the whole training stage, it runs 70 epoches and we do clustering with DBSCAN to generate new pseudo labels every two epoches. We did the experiments on two Tesla V100 GPU platform with 32G memory.

\subsection{Comparison with State-of-the-Arts on Purely USL Setting}
We compared our method with State-of-the-art methods under the purely USL setting on four commonly used datasets, Market1501, Duke, MSMT17 and PersonX, respectively. As illustrated in Table~\ref{table:Market1501-sta}, Table~\ref{table:Duke-sta}, Table~\ref{table:MSMT17-sta} and Table~\ref{table:PersonX-sta}, it shows that the proposed method outperforms all the latest unsupervised person Re-Id methods. Compared with the baseline method SPCL~\cite{zhu2020self}, the mAP of our proposed method surpasses the SPCL~\cite{zhu2020self} method by $8.6\%$, $5.2\%$, $5.5\%$ and $5.4\%$ on Martket1501~\cite{zheng2015scalable}, DukeMTMC~\cite{ristani2016performance}, MSMT17-v2~\cite{wei2018person}, and PersonX~\cite{sun2019dissecting} datasets respectively.
Since the proposed method improves the current unsupervised person Re-Id performance from different perspectives, we also formulate the latest un-published state-of-the-art method ``ClusterContrast" into the probability distillation framework which named ``ClusterContrast+Ours" in Table~\ref{table:Market1501-sta},\ref{table:Duke-sta},\ref{table:MSMT17-sta},\ref{table:PersonX-sta}, and further improves the mAP of the "ClusterContrast" method~\cite{dai2021cluster} by $1.9\%$, $0.7\%$, $2.0\%$ and $4.7$ on Martket1501~\cite{zheng2015scalable}, DukeMTMC~\cite{ristani2016performance}, MSMT17-v2~\cite{wei2018person}, and PersonX~\cite{sun2019dissecting} datasets respectively. Finally, our proposed method obtains a new state-of-the-art performances $84.5\%$, $73.5\%$, $24.6\%$ and $89.5\%$ mAP on all the four datasets, which surpass all published results for unsupervised person Re-Id tasks. Besides, the proposed method does not apply any extra parameters comparing with other state-of-the-art methods SPCL~\cite{zhu2020self} and "ClusterContrast"~\cite{dai2021cluster}.


\begin{table}[t]
\centering
\caption{Comparison with state-of-the-art methods on Market1501 Dataset under purely USL experiment setting.}\smallskip
\label{table:Market1501-sta}
\resizebox{0.9\columnwidth}{!}{
\begin{tabular}{c|c|c|ccc }
\toprule
\textbf{Methods} &\textbf{Reference} &\textbf{mAp} &\textbf{Top-1} &\textbf{Top-5}  &\textbf{Top-10} \\
   \hline
OIM~\cite{xiao2017joint} &CVPR2017 &14.0 &38.0 &58.0 &66.3 \\
LOMO~\cite{liao2015person} &CVPR2015 &8.0 &27.2 &41.6 &49.1 \\
BOW~\cite{zheng2015scalable} &ICCV2015 &14.8 &35.8 &52.4 &60.3 \\
BUC \cite{lin2019bottom}& AAAI2019 &38.3 &66.2 &79.6 &84.5 \\
UGA \cite{wu2019unsupervised} &CVPR2019 &70.3 &87.2  &-- &-- \\
\hline
SoftSim \cite{lin2020unsupervised}& CVPR2020 &37.8 &71.7 &83.8 &87.4 \\
TSSL \cite{wu2020tracklet}&AAAI2020 &43.3 &71.2 &-- &-- \\
MMCL \cite{wang2020unsupervised}&CVPR2020 &45.5 &80.3 &89.4 &92.3\\
JVTC \cite{li2020joint}&ECCV2020 &41.8 &72.9 &84.2 &88.7 \\
JVTC+\cite{li2020joint}& ECCV2020 &47.5 &79.5 &89.2 &91.9 \\
HCT \cite{zeng2020hierarchical} &CVPR2020 &56.4 &80.0 &91.6 &95.2 \\
CycAs \cite{wang2020cycas}&ECCV2020 &64.8 &84.8 &-- &--\\
MoCo~\cite{he2020momentum} &CVPR2020 &6.1 &12.8 &27.1 &35.7 \\
SPCL \cite{zhu2020self}  &NIPS2020 &73.1 &88.1 &95.1 &97.0 \\
\hline
JNTL \cite{Yang2021JointNoise} &CVPR2021 &61.7 &83.9 &92.3 &-- \\
JGCL \cite{chen2021joint} &CVPR2021 &66.8 &87.3 &93.5 &95.5 \\
IICS \cite{Yang2021JointNoise} &CVPR2021 &72.9 &89.5 &95.2 &97.0 \\
\hline
\textbf{Ours} & -- &\textbf{81.7} &\textbf{92.4} &\textbf{97.4} &\textbf{98.1}\\
\hline
ClusterContrast~\cite{dai2021cluster}& ArXiv2021 &82.6 &93.0 &97.0 &98.1 \\
\textbf{ClusterContrast+Ours}&-- &\textbf{84.5} &\textbf{93.5} &\textbf{97.6} &\textbf{98.6} \\
\bottomrule
\end{tabular}}
\end{table}

\begin{table}[t]
\centering
\caption{Comparison with state-of-the-art methods on DukeMTMC Dataset under purely USL experiment setting.}\smallskip
\label{table:Duke-sta}
\resizebox{0.9\columnwidth}{!}{
\begin{tabular}{c|c|c|ccc }
\toprule
\textbf{Methods} &\textbf{Reference} &\textbf{mAp} &\textbf{Top-1} &\textbf{Top-5}  &\textbf{Top-10} \\
   \hline
LOMO~\cite{liao2015person} &CVPR2015 &4.8 &12.3 &21.3 &26.6 \\
BOW~\cite{zheng2015scalable} &ICCV2015 &8.3 &17.1 &28.8 &34.9 \\
BUC \cite{lin2019bottom}& AAAI2019 &27.5 &47.4 &62.6 &68.4 \\
UGA \cite{wu2019unsupervised} &CVPR2019 &53.3 &75.0  &-- &-- \\
\hline
SoftSim \cite{lin2020unsupervised}& CVPR2020 &28.6 &52.5 &63.5 &68.9 \\
TSSL \cite{wu2020tracklet}&AAAI2020 &38.5 &62.2 &-- &-- \\
MMCL \cite{wang2020unsupervised}&CVPR2020 &40.2 &65.2 &75.9 &80.0\\
JVTC \cite{li2020joint}&ECCV2020 &42.2 &67.6 &78.0 &81.6 \\
JVTC+\cite{li2020joint}& ECCV2020 &50.7 &74.6 &82.9 &85.3 \\
HCT \cite{zeng2020hierarchical} &CVPR2020 &50.7 &69.6 &83.4 &87.4 \\
CycAs \cite{wang2020cycas}&ECCV2020 &60.1 &77.9 &-- &--\\
SPCL \cite{zhu2020self}  &NIPS2020 &65.3 &81.2 &90.3 &92.2 \\
\hline
JNTL \cite{Yang2021JointNoise} &CVPR2021 &53.8 &73.8 &84.2 &-- \\
JGCL \cite{chen2021joint} &CVPR2021 &62.8 &82.9 &87.1 &88.5 \\
IICS \cite{Yang2021JointNoise} &CVPR2021 &59.1 &76.9 &86.1 &89.8 \\
\hline
\textbf{Ours} & -- &\textbf{69.0} &\textbf{82.9} &\textbf{90.9} &\textbf{93.0}\\
\hline
ClusterContrast~\cite{dai2021cluster}& ArXiv2021 &72.8 &85.7 &92.0 &93.5 \\
\textbf{ClusterContrast+Ours}&-- &\textbf{73.5} &\textbf{85.4} &\textbf{92.2} &\textbf{94.5} \\
\bottomrule
\end{tabular}}
\end{table}

\begin{table}[t]
\centering
\caption{Comparison with state-of-the-art methods on MSMT17 Dataset under purely USL experiment setting.(*) indicates that the implementation is based on the authors' code.}\smallskip
\label{table:MSMT17-sta}
\resizebox{0.9\columnwidth}{!}{
\begin{tabular}{c|c|c|ccc }
\toprule
\textbf{Methods} &\textbf{Reference} &\textbf{mAp} &\textbf{Top-1} &\textbf{Top-5}  &\textbf{Top-10} \\
   \hline
TAUDL\cite{li2018unsupervised} &ECCV2018 &12.5 &28.4  &-- &-- \\
UGA \cite{wu2019unsupervised} &CVPR2019 &21.7 &49.5  &-- &-- \\
\hline
MoCo~\cite{he2020momentum} &CVPR2020 &1.6 &4.3 &-- &-- \\
MMCL \cite{wang2020unsupervised}&CVPR2020 &11.2 &35.4 &44.8 &49.8\\
JVTC \cite{li2020joint}&ECCV2020 &15.1 &39.0 &50.9 &56.8 \\
JVTC+\cite{li2020joint}& ECCV2020 &17.3 &43.1 &53.8 &59.4 \\
SPCL \cite{zhu2020self}  &NIPS2020 &19.1 &42.3 &55.6 &61.2 \\
\hline
JNTL \cite{Yang2021JointNoise} &CVPR2021 &15.5 &35.2 &48.3 &-- \\
JGCL \cite{chen2021joint} &CVPR2021 &21.3 &45.7 &58.6 &64.5 \\
IICS \cite{Yang2021JointNoise} &CVPR2021 &18.6 &45.7 &57.7 &62.8 \\
\hline
\textbf{Ours} & -- &\textbf{24.6} &\textbf{50.2} &\textbf{61.4} &\textbf{65.7}\\
\hline
ClusterContrast*~\cite{dai2021cluster}& ArXiv2021 &18.7 &40.5 &51.7 &56.9 \\
\textbf{ClusterContrast+Ours}&-- &\textbf{20.7} &\textbf{43.8} &\textbf{55.1} &\textbf{60.1} \\
\bottomrule
\end{tabular}}
\end{table}

 \begin{table}[t]
\centering
\caption{Comparison with state-of-the-art methods on PersonX Dataset under purely USL experiment setting.}\smallskip
\label{table:PersonX-sta}
\resizebox{0.9\columnwidth}{!}{
\begin{tabular}{c|c|c|ccc }
\toprule
\textbf{Methods} &\textbf{Reference} &\textbf{mAp} &\textbf{Top-1} &\textbf{Top-5}  &\textbf{Top-10} \\
   \hline
MMT~\cite{ge2020mutual} &NIPS2019 &78.9 &90.6 &96.8 &98.2 \\
SPCL \cite{zhu2020self}  &NIPS2020 &73.1 &88.1 &95.1 &97.0 \\

\hline
\textbf{Ours} & -- &\textbf{84.1} &\textbf{94.4} &\textbf{98.7} &\textbf{99.5}\\
\hline
ClusterContrast~\cite{dai2021cluster}& ArXiv2021 &84.8 &94.5 &98.4 &99.2 \\
\textbf{ClusterContrast+Ours}&-- &\textbf{89.5} &\textbf{95.6} &\textbf{98.7} &\textbf{99.4} \\
\bottomrule
\end{tabular}}
\end{table}

\subsection{Comparison with State-of-the-Arts on UDA Setting}

We also compare the proposed method with the state-of-the-art UDA person Re-Id methods, which utilize the labeled source domain datasets to assist the training on the unlabeled target datasets. Inspired by the state-of-the-art UDA person Re-Id method SPCL \cite{zhu2020self}, we further extend our method on the UDA setting. As presented in SPCL~\cite{zhu2020self}, we also consider the source domain class centroids from both the hybrid memory bank and the the training objectives. Thus the main differences between our method and the SPCL~\cite{zhu2020self} method under the UDA setting are described as follows: 1)In the training objective, we consider both the local batch contrastive the memory-based global contrastive learning. Besides, we always mine the hardest examples in each clustered group to represent the cluster for optimizing the objectives, without using the self-paced training strategy. While for the SPCL~\cite{zhu2020self} method, it optimizes the objectives using the memory-based contrastive learning with the class centroid representing the cluster in a self-paced manner; 2) The proposed method formulate the feature learning in the probability  distillation framework to utilize both the pseudo class labels and the self-supervised signals. The main similarity is that: we also use the labeled source domain dataset in the same manner as illustrated in SPCL~\cite{zhu2020self} and formulate them in the unified contrastive learning, which can make full use of all the source and target domain datasets, especially that the source domain dataset are well labeled.

As illustrated in Table~\ref{table:Duke-market1501-sta} and Table~\ref{table:Market1501-duke-sta}, we compare the proposed method with the state-of-the-art methods on two domain adaptation tasks, including $Duke\mapsto Market1501$ and $Market1501 \mapsto Duke$. It clearly shows that our proposed method outperforms all the other UDA methods, and we achieve $7.2\%$ and $2.1\%$ improvements in terms of mAP on the common $Duke\mapsto Market1501$ and $Market1501 \mapsto Duke$ tasks over state-of-the-art method SPCL~\cite{zhu2020self}, respectively. This greatly shows the robustness of the proposed method both on the purely unsupervised and UDA feature learning.

\begin{table}[t]
\centering
\caption{Comparison with state-of-the-arts on Market1501 Dataset under the UDA experiment setting.}\smallskip
\label{table:Market1501-duke-sta}
\resizebox{0.9\columnwidth}{!}{
\begin{tabular}{c|c|c|c|ccc }
\toprule
\textbf{Methods} &\textbf{Reference} &\textbf{Source} &\textbf{mAp} &\textbf{Top-1} &\textbf{Top-5}  &\textbf{Top-10} \\
   \hline
ECN~\cite{zhong2019invariance} &CVPR2019 &Duke &43.0 &75.1 &87.6 &91.6 \\
PDA~\cite{li2019cross} &ICCV2019 &Duke &47.6 &75.2 &86.3 &90.2 \\
CR-GAN~\cite{chen2019instance} &ICCV2019 &Duke  &54.0 &77.7 &89.7 &92.7 \\
SSG \cite{fu2019self}& ICCV2019 &Duke &58.3 &80.0 &90.0 &92.4 \\
CSCL \cite{wu2019unsupervised}& ICCV2019 &Duke &35.6 &64.7 &80.2 &85.6 \\
\hline
MEB-Net*\cite{zhai2020multiple} &ECCV2020 &Duke &71.9 &87.5 &95.2 &96.8 \\
ACT \cite{yang2020asymmetric} &AAAI2020 &Duke &60.6 &80.5  &-- &-- \\
MMCL \cite{wang2020unsupervised}&CVPR2020 &Duke &60.4 &84.4 &92.8 &95.0\\
DG-Net++ \cite{zou2020joint}&ECCV2020 &Duke  &61.7 &82.1 &90.2 &92.7 \\
JVTC \cite{li2020joint}&ECCV2020 &Duke  &61.1 &83.8 &93.0 &95.2 \\
JVTC+\cite{li2020joint}& ECCV2020 &Duke &67.2 &86.8 &95.2 &97.1 \\
ENC+\cite{zhong2020learning} &TPAMI2020 &Duke &63.8 &84.1 &92.8 &95.4 \\
CAIL~\cite{luo2020generalizing} &ECCV2020 &Duke &71.5 &88.1 &94.4 &96.2 \\
MMT~\cite{ge2020mutual} &ICLR2020 &Duke &71.2 &87.7 &94.9 &96.9 \\
NRMT~\cite{zhao2020unsupervised} &ECCV2020 &Duke &71.7 &87.8 &94.6 &96.5 \\
SPCL \cite{zhu2020self}  &NIPS2020 &Duke &76.7 &90.3 &96.2 &97.7 \\
\hline
JNTL \cite{Yang2021JointNoise} &CVPR2021 &Duke &76.5 &90.1 &-- &-- \\
JGCL \cite{chen2021joint} &CVPR2021 &Duke &75.4 &90.5 &96.2 &97.1 \\
\hline
\textbf{Ours} & -- &Duke &\textbf{83.9} &\textbf{93.6} &\textbf{97.5} &\textbf{98.3}\\
\bottomrule
\end{tabular}}
\end{table}

\begin{table}[t]
\centering
\caption{Comparison with state-of-the-arts on Duke Dataset  under the UDA experiment setting..}\smallskip
\label{table:Duke-market1501-sta}
\resizebox{0.9\columnwidth}{!}{
\begin{tabular}{c|c|c|c|ccc }
\toprule
\textbf{Methods} &\textbf{Reference} &\textbf{Source} &\textbf{mAp} &\textbf{Top-1} &\textbf{Top-5}  &\textbf{Top-10} \\
   \hline
ECN~\cite{zhong2019invariance} &CVPR2019 &Market &40.4 &63.3 &75.8 &80.4 \\
PDA~\cite{li2019cross} &ICCV2019 &Market &45.1 &63.2 &77.0 &82.5 \\
CR-GAN~\cite{chen2019instance} &ICCV2019 &Market  &48.6 &68.9 &80.2 &84.7 \\
SSG \cite{fu2019self}& ICCV2019 &Market &53.4 &73.0 &80.6 &83.2 \\
CSCL \cite{wu2019unsupervised}& ICCV2019 &Market &30.5 &51.5 &66.7 &71.7 \\
\hline
MEB-Net*\cite{zhai2020multiple} &ECCV2020 &Market &63.5 &77.2 &87.9 &91.3 \\
ACT \cite{yang2020asymmetric} &AAAI2020 &Market &54.5 &72.4  &-- &-- \\
MMCL \cite{wang2020unsupervised}&CVPR2020 &Market &51.4 &72.4 &82.9 &85.0\\
DG-Net++ \cite{zou2020joint}&ECCV2020 &Market  &63.8 &78.9 &87.8 &90.4 \\
JVTC \cite{li2020joint}&ECCV2020 &Market  &56.2 &75.0 &85.1 &88.2 \\
JVTC+\cite{li2020joint}& ECCV2020 &Market &66.5 &80.4 &89.9 &92.2 \\
ENC+\cite{zhong2020learning} &TPAMI2020 &Market &54.4 &74.0 &83.7 &87.4 \\
CAIL~\cite{luo2020generalizing} &ECCV2020 &Market &65.2 &79.5 &88.3 &91.4 \\
MMT~\cite{ge2020mutual} &ICLR2020 &Market &65.1 &78.0 &88.8 &92.5 \\
NRMT~\cite{zhao2020unsupervised} &ECCV2020 &Market &62.2 &77.8 &86.9 &89.5 \\
SPCL \cite{zhu2020self}  &NIPS2020 &Market &68.8 &82.9 &90.1 &92.5 \\
\hline
JNTL \cite{Yang2021JointNoise} &CVPR2021 &Market &65.0 &79.5 &-- &-- \\
JGCL \cite{chen2021joint} &CVPR2021 &Market &67.6 &81.9 &88.9 &90.6 \\
\hline
\textbf{Ours} & -- &Market &\textbf{70.9} &\textbf{83.5} &\textbf{90.8} &\textbf{93.3}\\
\bottomrule
\end{tabular}}
\end{table}

\subsection{Ablation Studies}
The proposed USL person Re-Id method contains two main novel ingredients: 1)The local-to-global memory-based contrastive learning with dynamic hardest example mining; 2)the proposed probability distillation framework to capture the self-supervised information. To reveal how each ingredient contributes to the performance improvements, we perform ablation study and implement different variants of the methods, then report the intermediate results of each component on Market1501 and Duke datasets.

\begin{table*}[t]\centering\small
\caption{Ablation study on the Market1501 and Duke datasets under the purely unsupervised person Re-Id task.}
\label{table:AblationStudyPUL}
\begin{tabular}{@{}cccc|cccc|cccc@{}}
\toprule

\multirow{2}{*}{\begin{tabular}[c]{@{}c@{}} \footnotesize{Centroids-Based}  \\ \footnotesize{Memory Contrastive} \end{tabular}} & \multirow{2}{*}{\begin{tabular}[c]{@{}c@{}} \footnotesize{Local-Batch} \\ \footnotesize{Contrastive}\end{tabular}} & \multirow{2}{*}{\begin{tabular}[c]{@{}c@{}} \footnotesize{Dynamic Hardest}\\  \footnotesize{example Mining}\end{tabular}} & \multirow{2}{*}{\begin{tabular}[c]{@{}c@{}} \footnotesize{Probability} \\ \footnotesize{Distillation}\end{tabular}} & \multicolumn{4}{c|}{Market1501}  &  &\multicolumn{3}{c}{Duke} \\ \cline{5-12}

& & &  &mAP     &Top-1     &Top-5     &Top-10  & mAP     &Top-1     &Top-5     &Top-10    \\ \hline

\ding{51}  & \ding{55}   & \ding{55}  & \ding{55}                & 76.3   & 90.1    & 96.1   & 97.3 & 61.9   &78.0    &87.1   &90.5  \\
\ding{51}  & \ding{51}   & \ding{55}  & \ding{55}                & 79.1   & 91.0    & 96.4  & 97.5  & 64.2  & 80.1    &88.2   &90.4 \\
\ding{51}  & \ding{51}   & \ding{51}  & \ding{55}                & 80.8   & 92.2   & 96.8  & 97.9  & 67.7   & 82.1    &90.1   & 92.6 \\
\ding{51}  & \ding{55}   & \ding{55}  & \ding{51}                & 78.1   & 90.6   & 96.5  & 97.5  & 64.8   & 80.4    & 88.2   & 90.5 \\
\ding{51}  & \ding{51}   & \ding{51}  & \ding{51}                & 81.7   & 92.4   & 97.4  & 98.1  & 69.0   & 82.9    & 90.9   & 93.0 \\ \bottomrule
\end{tabular}
\vspace{-5pt}
\end{table*}

\begin{table}[t]
\centering
\caption{ Ablation study on the Market1501 datasets under the UDA person Re-Id task.}\smallskip
\label{table:AlationUDA}
\resizebox{0.9\columnwidth}{!}{
\begin{tabular}{c|c|ccc }
\toprule
 Methods  &mAp &Top-1 &Top-5  &Top-10 \\
\hline
SPCL\cite{zhu2020self} &76.7 &90.3 &96.2 &97.7 \\
Baseline &80.0 &91.5 &96.8 &97.9 \\
$\mathcal{L}_{\text{L2G}}$    &82.5 &92.9 &97.1 &98.1 \\
Baseline+$\mathcal{L}_{\text{S}}$  &81.3 &92.2 &96.8 &97.7 \\
$\mathcal{L}_{\text{L2G}} + \gamma \mathcal{L}_{\text{S}}$   &83.9 &93.6 &97.5 &98.3 \\
\bottomrule
\end{tabular}}
\end{table}
As illustrated in Table~\ref{table:AblationStudyPUL}, the first line means that we use the estimated centroids in the contrastive learning framework as presented in SPCL~\cite{zhu2020self}, but we implement this without using the self-paced learning strategy. We can clearly see that our memory based global centroid contrastive learning can obtain $76.3\%$ mAP on Market1501 dataset, which performs better than the SPCL~\cite{zhu2020self} method ($72.6\%$ mAP). The second line is the proposed hybrid local-to-global cluster contrastive learning, we can clearly see that adding the local-batch contrastive learning could further improve the performance by $2.8\%$ and $2.3\%$ mAP on these two datasets respectively. The third line means that we propose to use the dynamic hardest example mining strategy to select one example in each group to replace the centroid in the contrastive learning objective as shown in Eq.~\ref{e:Center}, we can clearly see that such strategy can further improve the performance by $1.7\%$ and $3.5\%$ mAP on these two datasets respectively. The third line is to illustrate the effectiveness of the proposed probability distillation framework on top of our baseline method (line 1), we can clearly see that the probability distillation framework could well capture the self-supervised signals, which can improve the performances by $1.8\%$ and $2.7\%$ mAP on Market1501 and Duke datasets respectively. The last line is the results of our final USL algorithm. By comparing the last line with the third line, we can also see that the probability distillation framework can further improve the local-to-global cluster contrastive learning with dynamic hardest mining strategy by $0.9\%$ and $1.3\%$ mAP respectively.

Besides, we also illustrate the effectiveness of all the components in our proposed algorithm under the UDA setting on Market1501 dataset, as shown in Table~\ref{table:AlationUDA}. The first line in Table~\ref{table:AlationUDA} shows the performance of the SPCL~\cite{zhu2020self} algorithm, and the second line means that we use the estimated centroids in the contrastive learning framework as proposed in SPCL~\cite{zhu2020self}, but we implement this without using the self-paced learning strategy. It clearly shows that our proposed baseline method surpasses the SPCL method by $3.3\%$ mAP. The third line $\mathcal{L}_{\text{L2G}}$ is our proposed local-to-global memory based contrastive learning with the hybrid hardest example mining strategy, which improves the baseline method for $2.5\%$ mAP. The last two line illustrate the effectiveness of the proposed probability distillation framework $\mathcal{L}_{\text{S}}$ on top of the baseline method and the $\mathcal{L}_{\text{L2G}}$, the probability distillation module can further improve their corresponding baseline method by $1.3\%$ and $1.4\%$ mAP. Thus, we have illustrated the effectiveness of all the components in the proposed method under both the USL and UDA experiment settings.

\begin{table}[t]
\centering
\caption{ Ablation study of the probability distillation method on the Market1501 datasets.}\smallskip
\label{table:ProbabilityAblation}
\resizebox{0.9\columnwidth}{!}{
\begin{tabular}{c|c|ccc }
\toprule
 Methods  &mAp &Top-1 &Top-5  &Top-10 \\
\hline
$\mathcal{L}_{\text{L2G}}$  &80.8 &92.2 &96.8 &97.9 \\
$\mathcal{L}_{\text{L2G}}$+MeanTeacher-L2 &81.0 &92.4 &97.0 &98.1 \\
$\mathcal{L}_{\text{L2G}}+\lambda \mathcal{L}_{\text{S}}$  &81.7 &92.4 &97.4 &98.1 \\
\bottomrule
\end{tabular}}
\end{table}

\textbf{Probability Distillation} is one of the main module in the proposed framework to capture the self-supervised signals. To reveal some inner mechanism of the proposed distillation method, we thoroughly make detailed comparison with the traditional Mean-Teacher framework~\cite{cai2021exponential} on the Market1501 dataset under the purely USL setting. As illustrated in Table~\ref{table:ProbabilityAblation}, the first line is the baseline method $\mathcal{L}_{\text{L2G}}$ without using the probability distillation algorithm. The second line "$\mathcal{L}_{\text{L2G}}$+MeanTeacher-L2" is the traditional "Mean-Teacher" method~\cite{cai2021exponential}, where we have simply incorporated the proposed method $\mathcal{L}_{\text{S}}$ into the "Mean-Teacher" framework, and $L\_2$ regression loss is used to minimize the output features between the teacher and student networks. By comparing the results between $\mathcal{L}_{\text{L2G}}$ and $\mathcal{L}_{\text{L2G}}$+MeanTeacher-L2, the "$\mathcal{L}_{\text{L2G}}$+MeanTeacher-L2" algorithm obtains comparable results, and just improved the mAP by $0.2\%$ on Market1501 dataset. The third line is our proposed probability distillation method, where we have project all the image features onto the obtained cluster centroids and then get their corresponding pseudo labels. Finally, by using the proposed probability distillation method, we have further improved our baseline method $\mathcal{L}_{\text{L2G}}$ by another $0.9\%$ mAP, and we have got the new state-of-the-art results on all the four commonly used large-scale person re-Id datasets, under the purely USL and UDA settings.

\begin{table}[t]
\centering
\caption{ Ablation study of the parameter $\lambda$ on the Market1501 datasets.}\smallskip
\label{table:PersonX-sta}
\resizebox{0.9\columnwidth}{!}{
\begin{tabular}{c|c|ccc }
\toprule
 Parameter $\lambda$  &mAp &Top-1 &Top-5  &Top-10 \\
\hline
0 &80.8 &92.2 &96.8 &97.9 \\
0.1 &81.1 &92.2 &97.1 &98.1 \\
0.2 &\textbf{81.7} &92.4 &\textbf{97.4} &\textbf{98.1} \\
0.3   &81.6 &\textbf{92.5} &96.9 &98.1 \\
0.5 &80.9 &91.8 &96.9 &98.0 \\
1.0   &81.0 &92.4 &97.0 &98.1 \\
1.5  &79.9 &91.7 &96.6 &97.8 \\
\bottomrule
\end{tabular}}
\end{table}

\textbf{The hyper-parameter $\lambda$} in the final objective function is used as a tradeoff between the local-to-global cluster contrastive learning $\mathcal{L}_{\text{L2G}}$ and the probability distillation objective $\mathcal{L}_{\text{S}}$. As illustrated in Table~\ref{table:PersonX-sta}, we have made detailed analysis of the hyper-parameter $\lambda$ by varying its value from 0.0 to 1.5 on Market1501 dataset. We can clearly see that the proposed method obtains the best performances at $\lambda = 0.2$. By comparing all the results under different values of $\lambda$, the final results don't vibrate too much, which reveals that the proposed distillation framework is robust to such parameter and  benefits for network training.


\section{Discussion anf Conclusion}~\label{SEC:Conclusion}

In this paper, we propose the hybrid dynamic local-to-global cluster contrastive and probability distillation framework for unsupervised person re-Id task. Our algorithm surpasses almost all state-of-the-art methods on the four commonly used large-scale person re-Id datasets, namely Martket1501~\cite{zheng2015scalable}, DukeMTMC~\cite{ristani2016performance}, MSMT17-v2~\cite{wei2018person}, and PersonX~\cite{sun2019dissecting}. The proposed method formulates the unsupervised person Re-Id problem into an unified local-to-global dynamic learning and self-supervised probability regression framework. It can not only explore useful supervised information from the estimated pseudo labels by cluster contrast learning, but also make full use of the self-supervised signals of all the training images from both the instances' self-contrastive level and the probability regression perspective, in the memory-based non-parametric manner. We believe that our proposed unsupervised learning framework can be further extended to other unsupervised or semi-supervised learning tasks, and we will make much more detailed analysis and improvements of the proposed USL algorithm in the future.

\bibliographystyle{IEEEtran}
\bibliography{IEEEexample}


\end{document}